\documentclass[runningheads]{llncs}
\newcommand{\bftab}{\fontseries{b}\selectfont}
\usepackage[titlenumbered,ruled,linesnumbered, vlined]{algorithm2e}
\usepackage{amsmath} 
\usepackage{amssymb} 
\usepackage{graphicx}
\usepackage{float}
\usepackage{subcaption}
\usepackage{breakcites}
\usepackage{multirow}
\usepackage{hyperref}
\usepackage{xcolor}
\usepackage{import}
\makeatletter
\newcommand{\printfnsymbol}[1]{%
  \textsuperscript{\@fnsymbol{#1}}%
}

\begin{document}
\pagenumbering{arabic}

\title{Multimodal Meta-Learning for Time Series Regression}
%
%
\author{Sebastian Pineda Arango\thanks{Equal contribution.}\inst{1}, Felix Heinrich\printfnsymbol{1} \inst{2}, Kiran Madhusudhanan \inst{1}, Lars Schmidt-Thieme \inst{1}}
\authorrunning{S. Pineda Arango et al.}
%
\institute{University of Hildesheim, Hildesheim, Germany
 \\
\email{pineda@uni-hildesheim.de,\{kiranmadhusud,schmidt-thieme\}@ismll.uni-hildesheim.de} \and Volkswagen AG, Wolfsburg, Germany \\
\email{felix.heinrich1@volkswagen.de}
}
\maketitle              
\begin{abstract}
Recent work has shown the efficiency of deep learning models such as Fully Convolutional Networks (FCN) or Recurrent Neural Networks (RNN) to deal with Time Series Regression (TSR) problems. These models sometimes need a lot of data to be able to generalize, yet the time series are sometimes not long enough to be able to learn patterns. Therefore, it is important to make use of information across time series to improve learning. In this paper, we will explore the idea of using meta-learning for quickly adapting model parameters to new short-history time series by modifying the original idea of Model Agnostic Meta-Learning (MAML) \cite{finn2017model}. Moreover, based on prior work on multimodal MAML \cite{vuorio2019multimodal}, we propose a method for conditioning parameters of the model through an auxiliary network that encodes global information of the time series to extract meta-features. Finally, we apply the data to time series of different domains, such as pollution measurements, heart-rate sensors, and electrical battery data. We show empirically that our proposed meta-learning method learns TSR with few data fast and outperforms the baselines in 9 of 12 experiments.

\end{abstract}
\section{Introduction}

Time series regression is a common problem that appears when hidden variables should be inferred given a known multivariate time series. It finds applicability on a broad range of areas such as predicting heart-rate, pollution levels or state-of-charge of batteries. However, in order to train a model with high accuracy, a lot of data is needed. This sets some practical limitations, for instance, when a model is to be deployed on a new system with unknown conditions such as a new user or an older state of a battery, as new conditions cause a domain shift, to whom a model should be adapted based on few historical data. 

On the other hand, recent work on meta-learning for image classification has shown that it is possible to achieve fast adaptation when having few data \cite{finn2017model}. The core idea is to learn how to adapt the parameters efficiently by looking at many mutually-exclusive and diverse classification tasks \cite{rajendran2020meta}. Nevertheless, deriving a lot of tasks requires special task designs. In image classification, sampling classes and shuffling the labels have enabled these diverse tasks.

In time series regression problems, it is usual to have very long but few time series, as every time series is generated from a specific and small set of conditions (e.g. different subjects, machines, or cities). Therefore, applying powerful ideas from meta-learning is not trivial. A proof of it is the fact that there have been few works on how to apply those methods to time series regression, or even related problems such as time series forecasting.

The central idea of this work is to extend model-agnostic meta-learning (MAML) \cite{finn2017model} and multi-modal MAML (MMAML) \cite{vuorio2019multimodal} to time series regression. For that, we define the meta-learning problem as how to adapt fast to a new task given a set of known samples, namely \textit{support set}, so that we perform well on predicting the output channel for other samples that belong to the same task, or \textit{query set}.

We summarize the contributions from our paper as:

\begin{itemize}
    \item We propose a specific method for generating diverse tasks by assuming the real scenario of few but long time series at hand.
    \item It is the first work on how to extend MAML and MMAML to time series regression, and that demonstrates its improved performance over transfer learning.
    \item We show the utility of our ideas through empirical evaluations on three datasets and compare with different baselines by proposing an evaluation protocol that can be applied for future work on meta-learning for TSR.
\end{itemize}

\section{Related Work}

In recent years, there has been a lot of work on meta-learning applied on few-shot settings, specially in problems related to image classification and reinforcement learning \cite{finn2017model}\cite{vuorio2019multimodal}\cite{snell2017prototypical}\cite{oreshkin2018}. All of them share some commonalities, such as, an inner loop, or so-called \textit{base learner} that aims to  use the support set to adapt the model parameters, and an outer loop, or \textit{meta-learner}, that modifies the base-learner meta-parameters so that it learns faster. Sometimes the meta-learner is another model such as LSTM \cite{Sachin2017} or includes memory modules \cite{santoro2016meta}. Other approaches learn a metric function that allows finding fast a good prototype given a few samples of a new task \cite{snell2017prototypical}.

Nevertheless, among the landscape of all the methods for performing meta-learning, MAML stands out because it does not include additional parameters, but just aims to learn a good initialization so that a model achieves good performance after few gradient descent updates. Some methods, such as TADAM \cite{oreshkin2018} or Multimodal MAML (MMAML) \cite{vuorio2019multimodal} extend MAML by using additional networks for embedding the whole task and conditioning the parameters of the predictive model. Other methods simplify the optimization introduced in MAML by using first-order approximations to avoid computing Hessians \cite{nichol2018reptile}. 

It is possible to find previous work on meta-learning time series for specific problems. Lemke et al. \cite{lemke2010meta} propose a model that can learn rules on how to apply models on the Time Series Forecasting (TSF) problem. For that, they extract different features from the time series such as kurtosis and Lyapunov coefficient. Similarly, Talagala et al.\cite{talagala2018meta} train a Random Forest model that decides which is the best model to use on a new TSF problem. However, selecting the best model or creating rules for it does not yield a continuous search space. Given a specific support set, the set of possible models is discrete, and therefore, very limited. N-BEATS \cite{oreshkin2019n} is presented as a meta-learning option for zero-shot learning TSF that achieves good performance under unseen time series in \cite{oreshkin2020meta}. Although it was originally introduced as a model for purely TSF problems, the authors showed how the architecture resembles a meta-learner that adapts weights used for the final prediction. Nevertheless, the model itself does not provide explicitly a way to fine-tune using some samples (few-shot learning).

To deal with Time Series Classification (TSC) in few-shot settings, Narwariya et al.  \cite{narwariya2020meta} show how MAML can be adapted with minor modifications. They achieve better results compared to the baselines, where the Resnet achieved the closest performance results. Besides MAML, attention mechanisms have been used lately as an approach for leveraging the support set on text classification \cite{jiang2019attentive} and time series forecasting \cite{iwata2020few}.

\section{Multimodal Meta-Learning for TSR}

\subsection{Problem Definition}

We define generally meta-learning for time series regression as, given an ordered set $\mathcal{D}_j=\{(\mathbf{x}_n, y_n)\}_{n=1:N}$, the problem of learning a method that adapts the parameters $\theta$ of a regression model $f_{\theta}$ by using just $\mathcal{D}^{s}_j=\{(\mathbf{x}_n, y_n)\}_{n=1:Q}$ such that it performs well on $\mathcal{D}^{q}_j=\{(\mathbf{x}_n, y_n)\}_{n=Q:N}$. Where $\mathbf{x}_n \in \mathrm{R}^{L \times C}$ denotes a time window with labels $y_n \in \mathrm{R}$, whereas $\mathcal{D}^s_j$ and $\mathcal{D}^q_j$ are the support and query sets, respectively. The union of both sets, $\mathcal{D}_j=\{(\mathbf{x}_n, y_n)\}_{n=1:N}$, is a time series regression (TSR) task, a definition inspired by the concept of task in meta-learning applied to image classification \cite{finn2017model}\cite{snell2017prototypical}. Every fixed-length window is intended to be the input for the regressor.

Formally, the optimization objective can be defined as

\begin{equation}
\min_{\phi, \theta} \sum_j \mathcal{L}_j(\mathcal{D}^q_j, f_{\theta^*})
\label{eq:ml_for_tsr_problem}
\end{equation}
where $\theta^*= \mathcal{U}(\theta, \mathcal{D}^{s}_j, \phi)$ is a learner or an update rule for the parameters $\theta$ that depends on the support set $\mathcal{D}^s_j$ and the meta-parameters $\phi$.  The task loss $\mathcal{L}_j$ measures the performance of the predicted labels accounting for all the labeled windows $\mathcal{L}_j(\mathcal{D}^q_j, f_{\theta |\phi}) = \sum_n \mathcal{L}(y_n, f_{\theta }(\mathbf{x}_n))$. Since this is a regression problem, the loss can be, for instance, the Mean Squared Error (MSE) or the Mean Absolute Error (MAE).

When applied to a new TSR task, the learner uses the update rule to estimate a new set of parameters $\theta^*$ by just using Q samples (or time windows in this context). Q is typically small, therefore training a network from scratch with $\mathcal{D}^s_j $ is not possible without overfitting. However, in order to tackle this problem, there is normally a small set of long multivariate time series (including the target channel) $\mathcal{S}$ available for training such that $\mathcal{S}= \{(\mathbf{S}_i, Y_i)|\space \mathbf{S}_i \in \mathrm{R}^{L_i \times C}, Y_i \in \mathrm{R}^{L_i}, i=1, ..., M \} $, where $\mathbf{S}_i$ denotes the input channels and $Y_i$, the output or target channel to be predicted.

\subsection{Meta-Windows: Redesigning Tasks for TSR }
\label{section:meta-windows}
Independently from the approach to solve the problem formulated in equation \ref{eq:ml_for_tsr_problem}, it is necessary to have a lot of tasks available for training, this is also the case for meta-learning applied to image classification. Nevertheless, the situation is usually to have long but few multivariate time series. In this subsection, we introduce a redesign of this setting to overcome this challenge. 

Long time series are difficult to feed into a model, therefore the common approach consists in creating smaller fixed-length windows through a windows generation process that uses a rolling window, $\mathcal{W}_{\delta, k}(\cdot)$, where $\delta$ denotes the window size and $k$ is the step size for the windows generation. Given a long multivariate time series with target channel, $(\mathbf{S}, Y) \in \mathcal{S}$, we generate the set of labeled windows $\mathcal{D}$ such that $\mathcal{D}=\mathcal{W}_{\delta, k}(\mathbf{S}, Y)=\{(\textbf{x}_n, y_n), \textbf{x}_n = \textbf{S}_{ (j\cdot k) : (j \cdot k+\delta)}, y_n = Y_{{(j \cdot k +\delta)}}, j=1,...,L-\delta\}$, where $L$ is the length of the multivariate time series and the lower indexing on $S$ and $Y$ refers to the time axis of the time series.

The methods introduced in this paper leverage these windows by grouping them in \textbf{meta-windows}. All meta-windows contain the same number of labeled windows (denoted as $l$), whereas every window belongs to only one window. The algorithm \ref{alg:meta-windows_generation} explains how the meta-windows are generated from long time series.

 \begin{figure} []
\centering
\includegraphics[width=8cm]{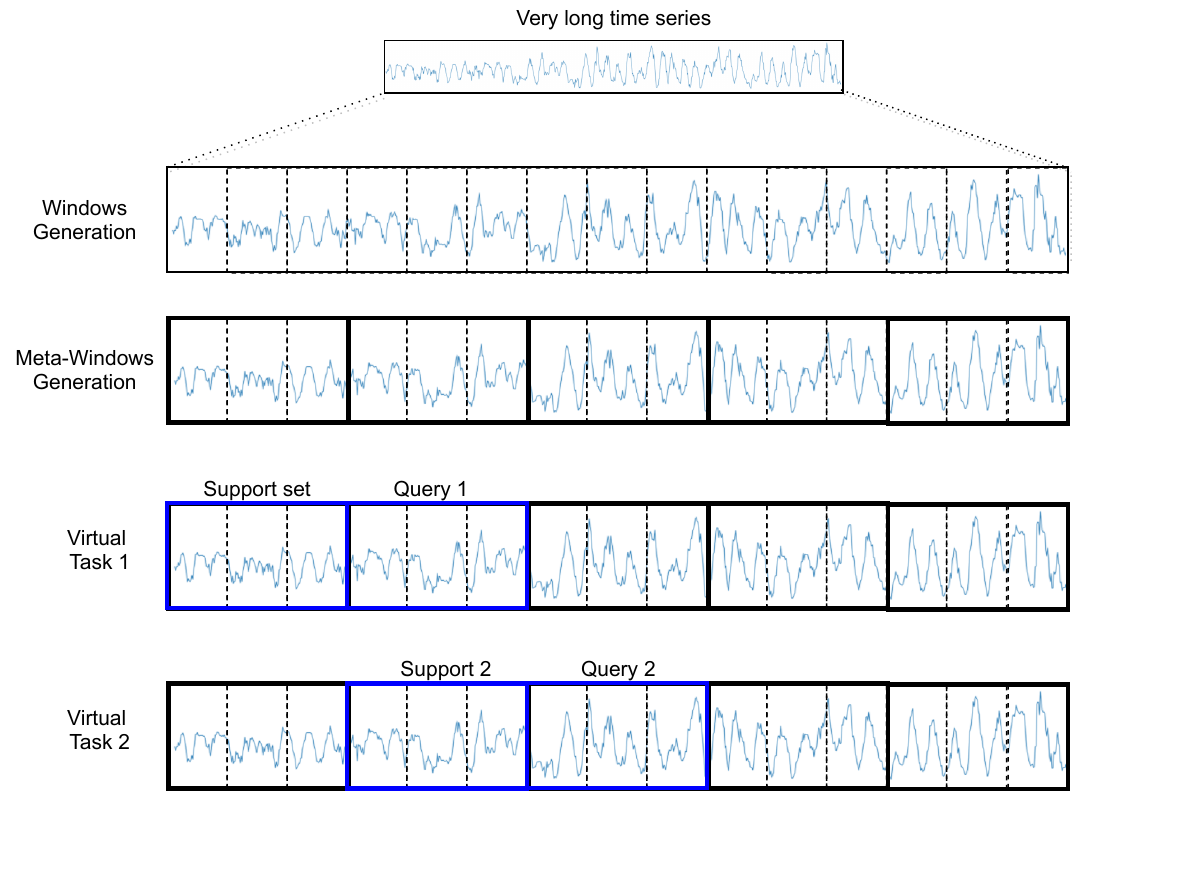}
\caption{Task design for a univariate time series, also applicable to multivariate time series. We omit the other channels and the respective target associated to every window for the sake of simplicity. }
\label{fig:task_design}
\end{figure}

If the time series are not periodic, it is possible to assume that the windows are very correlated to other temporally close windows, but are less correlated to the temporally far samples. In fact, this can be supported by looking at the monotonically decreasing auto-correlation diagram of $Y_i$ (see supplementary material \footnote{Accessible in \url{https://www.dropbox.com/s/tuzs6l8zy9zyon9/AALTD_21_MMAML_TSR_Supplementary.pdf?dl=0}}). This results in a meta-window $\mathcal{T}_t \in \mathcal{T}$ that is correlated with its  neighbor $\mathcal{T}_{t+1}$, but approximately uncorrelated with other meta-windows.

Based on the above-mentioned assumption on temporally correlation, we redesign a task for TSR such that two continuous windows are considered belonging to the same task. It means, after the defined problem in equation \ref{eq:ml_for_tsr_problem}, we can generate a lot of tasks by setting a sampled meta-window $\mathcal{T}_t \sim \mathcal{T}$  as the support set $\mathcal{D}^s_j$
 and the next one $\mathcal{T}_{t+1}$ as the query set. Also, due to the decreased correlation with temporally-far meta-windows, this design guarantees certain level of diversity. Moreover, including meta-windows generated from different long time series may increase this task diversity. Since the generated pair of sets support-query aim to "simulate" a new TSR task, we refer to them as \textit{virtual tasks}. The figure \ref{fig:task_design} illustrates this procedure.
 
 \subsection{MAML for TSR}

In this section, we formally define our proposed algorithm MAML for TSR. We denote $\mathcal{T}_t$ as the $t$-th meta-window, sampled from a distribution $p(\mathcal{T})$. We assume that the meta-windows were generated from a set of long multivariate time series coming from different instances but with the same semantics among the channels (i.e. three accelerometer measurements from different subjects). 

Given that the meta-windows, coming from the same long time series, are temporally ordered through a temporal index $t$, we want to use the meta-window 
 
\begin{algorithm}[h]

\SetAlgoLined

\SetKwInOut{Input}{Input}
\Input{Long multi-variate time series with target channel $\mathcal{S} = \{(\mathbf{S}_i, Y_i)|\space \mathbf{S}_i\in \mathrm{R}^{L_i\times C}, Y_i \in \mathrm{R}^{L_i}, i=1, ..., N \}  $, a rolling window $\mathcal{W}_{\delta, k}(\cdot)$}
\Input{Hyper-parameters  $l, \delta, k:$ meta-window length, window size and step size}
Initialize ordered set of meta-windows $\mathcal{T}$\\
\textbf{for all} $ (\mathbf{S}_i, Y_i) \in \mathcal{S}$:\\
$\quad$ Create ordered set of tuples ${\mathcal{D}} = \mathcal{W}_{\delta, k} (\mathbf{S}_i, Y_i) =  \{(\textbf{x}_n, y_n) \}$ \\
$\quad$ \textbf{for} $n=1,..., \lfloor{\frac{{|\mathcal{D}|}}{l}} \rfloor $:\\
$\quad \quad$ Append $\{{\mathcal{D}}_{n\cdot l:(n+1)\cdot l} \}$ to $\mathcal{T} $  \\
$\quad$ \textbf{end for}\\
\textbf{end for} \\
\caption{Meta-windows generation.}
\label{alg:meta-windows_generation}

\end{algorithm}

\begin{algorithm}[h]

\SetAlgoLined

\SetKwInOut{Input}{Input}
\Input{ $p(\mathcal{T}):$ distribution over meta-windows, with indexed, and temporally ordered meta-windows $\mathcal{T}_1, \mathcal{T}_2, ...$ }
\Input{$\alpha$: learning rate, $\beta$: meta-learning rate, $\mathcal{L}_{\mathcal{T}}$: task loss} 

randomly initialize $\theta$ \\
\textbf{while} not done \textbf{do}   \\
$\quad$ Sample batch of meta-windows $\mathcal{T} \sim p(\mathcal{T})$ \\
$\quad$ \textbf{for all $\mathcal{T}_{t} \in \mathcal{T}$  do}:\\
$\quad \quad$ Set windows $\mathcal{T}_{t}=\left\{\mathbf{x}^{(t)}_{i}, y^{(t)}_{i}\right\}$  \\
$\quad \quad$ Evaluate $\nabla_{\theta} \mathcal{L}_{\mathcal{T}_{t}}\left(f_{\theta}\right)$ using $\mathcal{T}_t$ and $\mathcal{L}_{\mathcal{T}_{t}}$ \\
$\quad \quad $  Compute parameter updates : ~
$\theta_{t}^{\prime}=\theta-\alpha \nabla_{\theta} \mathcal{L}_{\mathcal{T}_{t}}\left(f_{\theta}\right)$ \\
$\quad \quad $
Save windows $\mathcal{T}_{t+1}=\left\{\mathbf{x}^{(t+1)}_{j}, {y}^{(t+1)}_{j}\right\}$ for
meta-update \\
$\quad $ \textbf{ end for} \\
$\quad$Update $\theta \leftarrow \theta-\beta \nabla_{\theta} \sum_{\mathcal{T}_{t} \in \mathcal{T}} \mathcal{L}_{\mathcal{T}_{t+1}}\left(f_{\theta_{t}^{\prime}}\right)$  using dataset $\mathcal{T}_{t+1}$ and a given model loss $\mathcal{L}_{\mathcal{T}_{t+1}}$.\\
\textbf{end while} \\
\caption{MAML for TSR}
\label{alg:maml_for_tsr}

\end{algorithm}

\noindent $\mathcal{T}_t$ as the support set, while setting the subsequent meta-window $\mathcal{T}_{t+1}$ as the query. The idea of MAML for TSR is, then, to find an initial set of parameters $\theta^*$ such that a gradient descent optimizer adapts the model $f_{\theta}$ in one step (or few steps) to a new domain by using just a meta-window $\mathcal{T}_t$. This optimization objective can be expressed formally as:

\begin{equation}
\centering
\begin{split}
   \min _{ \theta} \sum_{\mathbf{\mathcal{T}}_{t} \sim p(\mathcal{T})} \mathcal{L}_{\mathbf{\mathcal{T}}_{t+1}}\left(f_{ \theta-\alpha \nabla_{\theta} \mathcal{L}_{\mathcal{T}_{t}}\left(f_{\theta}\right)}\right) 
  \\
   =\min_{ \theta }\sum_{\mathcal{T}_t \sim p(\mathcal{T}) }  \sum_{{ (\mathbf{x}_j, {y}_j) \in \mathcal{T}_{t+1}  }} ||{y}_j - f_{ \theta-\alpha \nabla_{ \theta}\mathcal{L}_{\mathcal{T}_{t}}(f_{ \theta})}(\mathbf{x}_j)  ||_1
\end{split}
\label{eq:maml_for_tsr}
\end{equation}

which is based on the formulation of MAML by Finn et al. \cite{finn2017model} for fast adaptation in classification with neural networks. The second line reformulates the loss by using MAE as the task loss $\mathcal{L}_{\mathcal{T}}$. In the algorithm \ref{alg:maml_for_tsr}, we detail the process for optimizing the proposed loss function in equation \ref{eq:maml_for_tsr}.
\newline

\subsection{MMAML for TSR}

In this section, we introduce the multi-modal model-agnostic meta-learning (MMAML) for TSR which draws inspiration from \cite{vuorio2019multimodal}. This approach takes a \textbf{modulation network} that changes the parameters of a \textbf{task network} which makes the final prediction. The parameters are modulated according to meta-task information that is extracted by the modulation network. Therefore, the modulation network is a feature extractor at the task level (or meta-window level), whereas the task network processes single windows. The extracted information from the meta-windows is useful for the fast adaptation. In the supplementary material, we show that the embeddings of the meta-windows are forming groups with the other ones coming form the same long time series.

Encoding task information means in our current work to embed the support set, which is a meta-window. Vuorio et al. \cite{vuorio2019multimodal} use relational networks to process images and targets belonging to the support set. However, as we are interested in embedding meta-windows (time series), we need to perform the task encoding differently.

\begin{algorithm}[]
\SetAlgoLined
\SetKwFunction{FSummarize}{Summarize}
\SetKwFunction{FFilm}{FiLM}
\SetKwInOut{Input}{Input}
\Input{ $p(\mathcal{T}):$ distribution over meta-windows, with indexed, and temporally ordered meta-windows $\mathcal{T}_1, \mathcal{T}_2, ...$ }
\Input{ $\alpha, \beta:$ step size hyper-parameters, $\lambda$ : weight for the variational loss}

randomly initialize $\omega = \{\theta, \theta_{dec}, \theta_{enc}, \theta_{gen}, \theta_{ext}\}$  \\
\textbf{while} not done \textbf{do}   \\
$\quad$ Sample batch of meta-windows $\mathcal{T} \sim p(\mathcal{T})$ \\
$\quad$ \textbf{for all $\mathcal{T}_{t} \in \mathcal{T}$  do}:\\
$\quad \quad$ Set windows $\mathcal{T}_{t}=\left\{\mathbf{x}_{i}, {y}_{i}\right\}$ \\
$\quad \quad$ Get meta-window summary: $\mathcal{T}^{\prime}_t  = \FSummarize(\mathcal{T}_t)$ \\
$\quad \quad$ Infer embedding: $z = h_{\theta_{enc}}(\mathcal{T}^{\prime}_t)$ \\
$\quad \quad$ Generate parameters: $\rho = h_{\theta_{gen}}(z)$ \\
$\quad \quad$ Modulate last layer parameters: $\hat{\theta} = \FFilm(\theta| \rho)$ \\
$\quad \quad$ Evaluate $\nabla_{\theta} \mathcal{L}_{\mathcal{T}_{t}}\left(f_{\hat{\theta}}\right)$ using $\mathcal{T}_{t}$ and $\mathcal{L}_{\mathcal{T}_{t}}$ \\
$\quad \quad $  Compute parameter updates: ~
$\theta_{t}^{\prime}=\theta-\alpha \nabla_{\theta} \mathcal{L}_{\mathcal{T}_{t}}\left(f_{\hat{\theta}}\right)$ \\
$\quad \quad $
Save meta-window $\mathcal{T}_{t+1}=\left\{\mathbf{x}_{j}, {y}_{j}\right\}$ for meta-update \\
$\quad$\textbf{end for} \\
$\quad$Reconstruct meta-windows $\mathcal{T}_t \in \mathcal{T}$: $\hat{\mathcal{T}_t}=h_{\theta_{dec}}(h_{\theta_{enc}}(\mathcal{T}^{\prime}_t))$  \\
$\quad$Update $\omega \leftarrow\omega-\beta \nabla_{\omega} \sum_{\mathcal{T}_{t} \in \mathcal{T}} \left( \mathcal{L}_{\mathcal{T}_{t+1}}\left(f_{\theta_{t}^{\prime}|\rho}\right) + \lambda \mathcal{L}_{VAE}({\mathcal{T}_t}^{\prime}) \right)$\\

\textbf{end while} 
\caption{MMAML for TSR}
\label{alg:mmaml_for_tsr}

\end{algorithm}
\subsubsection{\textbf{Meta-Windows Encodings}} As explained before, the support set (task information) is a meta-window $\mathcal{T}= \{{(\mathbf{x}_i, {y}_i}), i =1,..,l\}$ and is originated from a long time series $(\mathbf{S}, Y)$. As a way of simplifying the meta-window, so that it can be input to the modulation network without redundant information and by avoiding a huge overhead in training, we propose to summarize it by concatenating the first sample of every window (and every channel) belonging to the meta-window. An additional channel is created after concatenating similarly the respective target signal. Therefore, the summarization is performing a downsampling of the meta-window. After summarizing the support set, we obtain a multivariate-time series (MTS), which can be encoded through any representation learning algorithm. For learning this latent representation, we use a variational recurrent auto-encoder (VRAE)\cite{fabius2014variational}.

\subsubsection{\textbf{Modulation and task network}} We introduce the modulation network to be applied in our work, which comprises three sub-modules:

\begin{itemize}
\item  \textbf{Encoder}. It embeds the input (summarized meta-window $\mathcal{T} ^{\prime}$) to a latent representation, such that $z = h_{\theta_{enc}}(\mathcal{T  ^{\prime}})$ using variational bayes. 
\item  \textbf{Decoder}. It reconstructs the input, formally $\hat{\mathcal{T}} = h_{\theta_{dec}} (z)$, aiming to minimize the reconstruction loss $\|\mathcal{T ^{\prime}}- \hat{\mathcal{T}}\|^2$.
\item  \textbf{Generator}. It outputs the parameters $\rho$ that modify the parameters $\theta$ from the main network through FiLM layers \cite{perez2018}. Formally, $\rho = h_{\theta_{gen}}(z)$.
\end{itemize}

The task network is composed of a feature extractor $\phi_{\theta_{ext}}$ and a last (linear) layer $\theta$. The final output for an input \textbf{x}, given the output of the generator $\rho$, is then computed as:

\begin{equation}
\hat{y} = f_{\theta| \rho} (\textbf{x})= {FiLM}(\theta | \rho)^T \phi_{\theta_{ext}}(\textbf{x})
\label{eq:mmaml_output}
\end{equation}
Here we have a set of four types of parameters $\omega = \{\theta, \theta_{dec}, \theta_{enc}, \theta_{gen}, \theta_{ext}\}$, but $\theta$ being the most interesting ones as they are the only fine-tuned parameters for fast adaptation in this work. In general, however, the parameters of the feature extractor can be also included in the adaptation. 

The loss for MAML (equation \ref{eq:maml_for_tsr}) can be extended easily to a formulation that includes the modulation network.
\begin{equation}
     \sum_{{ (\mathbf{x}_j, {y}_j) \in \mathcal{T}_{t+1}  }} ||{y}_j - f_{ \theta-\alpha \nabla_{ \theta}\mathcal{L}_{\mathcal{T}_{t}}(f_{ \theta | \rho}) | \rho}(\mathbf{x}_j)  ||_1 + \lambda \cdot \mathcal{L}_{VAE}(\mathcal{T}^{\prime}_t)
\label{eq:mmaml_for_tsr}
\end{equation}
We have included the variational loss that can be re-phrased in our context, by denoting $\mathcal{T}^{\prime}$ as the meta-window summary, in the following way:
\begin{equation}
\mathcal{L}_{VAE}(\mathcal{T}^{\prime}_t) =\| \mathcal{T}_t ^{\prime}- \hat{\mathcal{T}_t}\|^{2} + \mathcal{K} \mathcal{L}[\mathcal{N}(\mu( \mathcal{T}_t^{\prime}), \Sigma(\mathcal{T}_t ^{\prime})) \| \mathcal{N}(0, I)],
\label{eq:vae_loss}
\end{equation}

where $ \mathcal{K} \mathcal{L}$ is the Kullback-Leiber divergence, $I$ is an identity matrix and $\mu(\cdot), \Sigma(\cdot)$ are mean and covariance functions respectively, which are modelled by the encoder.

\begin{figure} []
\centering
\includegraphics[width=8cm]{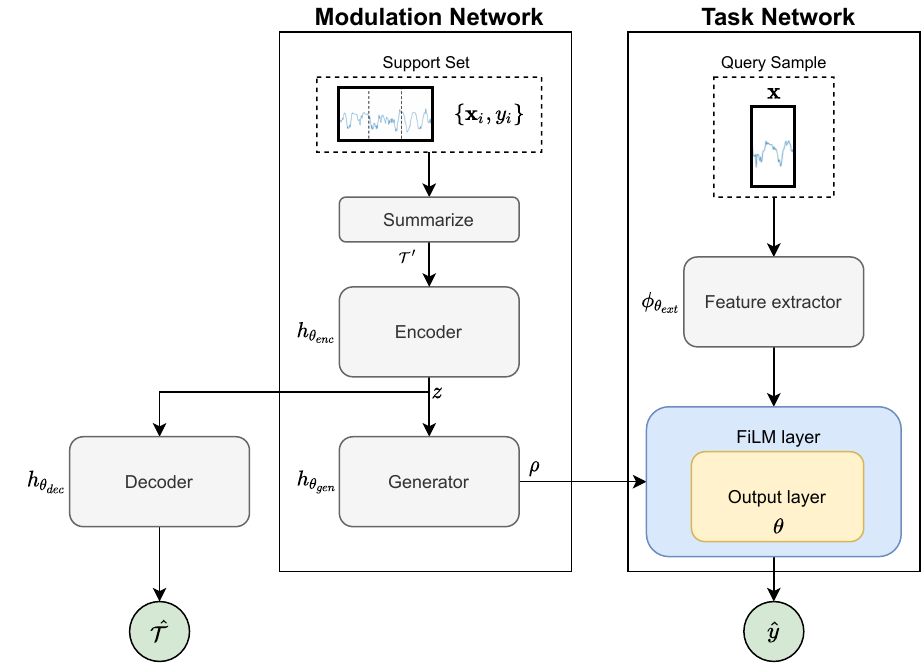}
\caption{MMAML for TSR architecture (based on \cite{vuorio2019multimodal}).}
\label{fig:mmaml_architecture}
\end{figure}
The algorithm \ref{alg:mmaml_for_tsr} introduces MMAML for TSR. It comprises an outer loop which modifies all the involved parameters $\omega$, and an inner-loop that just involves the parameters meant to be updated during the fast adaptation, $\theta$. 

The figure \ref{fig:mmaml_architecture} illustrates the different modules. Note that there are two networks: the modulation and task network. The decoder is considered outside the task network as it is not used in inference time. More importantly, the input of the task network is just a window, while the input for the modulation network is a summarized meta-window.

\section{Experiments}
\label{section:experiments}

In this section, we test our proposed ideas empirically to assess how much improvement meta-learning can bring compared to other approaches. Furthermore, we are interested in analyzing the performance of the models after several gradient steps and in long-term, in other words, in time horizons very far from the support meta-window that has been used for the adaptation.

\subsection{Datasets}

We perform the experiments on three different datasets which belong to different use cases.

The \textbf{air pollution dataset} (whose original name is \textit{2.5 Data of Chinese Cities Dataset}\footnote{\url{https://archive.ics.uci.edu/ml/datasets/PM2.5+Data+of+Five+Chinese+Cities}}, and from now on referred as \textbf{POLLUTION}) contains the PM 2.5 data from Beijing, Shanghai, Guangzhou, Chengdu and Shenyang, which includes as input channels meteoreological variables, whereas the target channel corresponds to the PM2.5 particles concentrations. The time period of the data spans six years, between Jan 1st, 2010 to Dec. 31st, 2015. In this dataset, every city corresponds to a very long time series that is the source for the meta-window generation.

The \textbf{heart rate dataset} (whose original name is \textit{PPG-DaLiA Dataset}\footnote{\url{https://archive.ics.uci.edu/ml/datasets/PPG-DaLiA}}, from now on referred as \textbf{HR}) is to be used for PPG-based heart rate estimation. It includes physiological and motion  variables, recorded from wrist- and chest-worn devices, on 15 subjects performing different activities, under real-life conditions. We consider every subject as a very-long time series. A careful pre-processing is necessary to synchronize the variables samples, as they have different sampling frequencies, and to generate the ground-truth heart rate. More about this is explained in the supplementary material.

The \textbf{battery dataset} is provided by Volkgswagen AG, and from now on is referred as \textbf{BATTERY}. It is intended to predict the current voltage, given past and current values of the current, temperature and charge. The data is divided in eleven different folders, where every folder corresponds to a specific battery age. Hence, the folders span aging battery information from 0 to 10 years. Every folder contains different files (every file is considered a very long time series, having 96 in total) that matches a specific driving cycle with different patterns of speed, from full charge till the discharge of the battery.

Finally, the windows size used for POLLUTION, HR and BATTERY are respectively 5, 32 and 20. An explanation for this decision is given in the supplementary material.

\subsection{Baselines}
We compare our proposed methods with the following models:

\begin{itemize}
    \item {\textbf{Target Mean}} We predict for all the query samples the mean of the target from the support set such that $\hat{y}= \frac{1}{l}\sum_{(\mathbf{x}_i, y_i) \in \mathcal{T}_{t}} y_i$. However, this corresponds to an unreal scenario, as we are assuming that the target channel is not available in inference time, but just in fast adaptation. 
    \newline
    \item \textbf{Resnet}  We used the implementation of \cite{tan2020time} keeping the same architecture (filter size and number of filters). The number of parameters of the final model is about 500.000. This model is considered, because it has shown competitive results in time series regression \cite{tan2020time}. Before adapting the model to a new (virtual) task, we pre-train the network on the raw meta-training dataset using standard training. Then, we apply transfer learning, where we freeze the whole network parameters but the last layer (128 parameters). \newline
    \item {\textbf{VRADA}} It \cite{sanjay2017} combines two networks: a VRNN as proposed by \cite{chung2015recurrent} and DANN \cite{ganin2016domain}, which includes a domain classifier and a regressor. The VRNN is a one-layer LSTM, with hidden dimension and the latent dimension equal to 100 (as in the original paper). The domain classifier comprises two fully connected layers with 100 and 50 neurons respectively, with output layer equal to the \textit{number of different classes} (which in this case we assume are each of the original long time series in $\mathcal{S}$), whereas the label predictor is a regressor with similar architecture but with output layer equal to 1. The hidden layers use DropOut (drop-out probability = 0.5), Batch Normalization and ReLU. The number of total parameters for VRADA is around 236.000. It is considered as a baseline since its architecture allows to extract domain-invariant features.
   \newline
   \item \textbf{LSTM} \cite{hochreiter1997long} We applied an architecture with 120 neurons, 2 layers and a linear output layer. The total number of parameters, varying according to every dataset, is around 180.000. Although this is also the model used for applying MAML and MMAML, we use as baseline a standard fine-tuning, after pre-training a network without meta-learning. 
\end{itemize}

\subsection{Experimental setup}

Given a set of few very long-time series, we split them in three disjunctive sets: training, validation and testing. 60\% of the very-long time series are assigned to training, 20\% to validation and 20\% to test. For every split, we run a rolling window that generates a set of labeled windows. Subsequently, meta-windows are generated by applying the algorithm \ref{alg:meta-windows_generation} on every split, thus originating the datasets splits used on meta-learning: meta-training, meta-validation and meta-testing respectively\footnote{We provide the created meta-windows for the splits of POLLUTION and HR as pickled numpy objects in \url{https://www.dropbox.com/sh/yds6v1uok3bjydn/AAC5GRWw0F3clopRlk00Smvza?dl=0}}.

MAML and MMAML use the meta-training and meta-validation datasets, whereas the baselines are trained using a training and validation datasets. After training (or meta-training), the final parameters are available to be used as initialization for fine-tuning. How good the models after fine-tuning are is evaluated using a meta-testing protocol, where every experiment is run five times to report the mean and the 95\% confidence interval.

\begin{figure} []
\centering
\includegraphics[width=10cm]{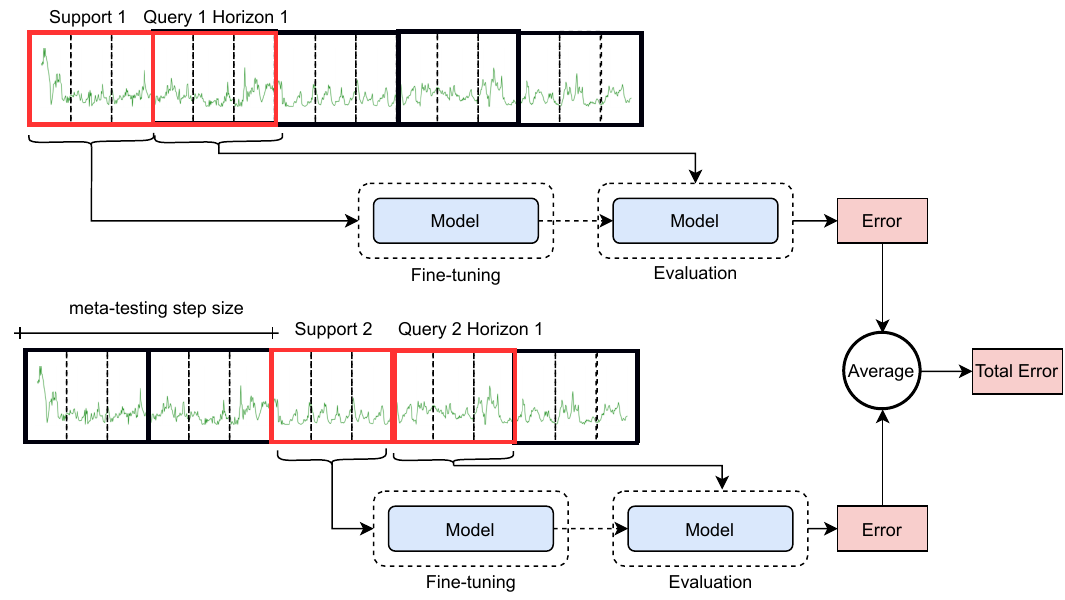}
\caption{Meta-testing Protocol. This procedure enables a lot of tasks during evaluation.}
\label{fig:meta-testing}
\end{figure}

The meta-testing follows the same procedure for the baselines and MAML/MMAML. We draw a meta-window as the support set, and use it to adapt (or fine-tune) the model, while the following meta-windows (up to a given horizon $H$) are used as query set. By doing this iteratively, many virtual tasks are available for the meta-testing, as simply sliding over the ordered set of meta-windows with a given step size (meta-testing step size) gives different virtual tasks. For instance, figure \ref{fig:meta-testing} depicts two virtual tasks separated by a meta-testing step size equals to two. At the end, the total error is computed as the average error over all the queries. To follow the proposed meta-testing procedure, two parameters must be provided: the meta-testing step size and horizon.

Specifically, we experiment fine-tuning with 1 and 10 gradient updates of the \textbf{last layer} parameters, except for VRADA, where all the parameters of its regressor module are updated. For the baselines, we also experimented with updates including weight decay. Note that MAML and MMAML also consider this number of updates in the inner-loop during meta-training, not only in the meta-testing. Additionally, in order to assess how good was the adaptation to the new virtual task on the long term, we evaluate on a query set that includes then 10 following meta-windows (10 horizons). The meta-testing step-size is set to $\lfloor \frac{\textit{M}}{100} \rfloor$, where $M$ is the number of meta-windows generated from the long time series $\mathcal{S}_i = (\mathbf{S}_i, Y_i) \in \mathcal{S}$. Additionally, we experiment with a meta-window length equal to 50 ($l=50$).

The training phase of the baselines uses the training set for optimizing the parameters of the whole network by using mean absolute error (MAE) as loss $\mathcal{L}_{\mathcal{T}_t}$ and the labeled windows. The number of iterations was limited by the early stopping criteria, and using the error in the validation/meta-validation set as a reference. If the error in the validation set does not decrease within a given number of iterations (so-called \textbf{patience}), then the training stops. A \textbf{training epoch} is considered to be the group of updates after passing over the whole training set. 

Similarly, the training of MAML/MMAML, here referenced as meta-training, follows the algorithm \ref{alg:maml_for_tsr} and \ref{alg:mmaml_for_tsr} respectively, only adapting the last layer parameters. Early stopping was also applied, so that it stops when the meta-validation error starts to increase after some epochs. In this context, a \textbf{meta-training epoch} is one outer-loop iteration in the algorithm \ref{alg:mmaml_for_tsr}, therefore it does not compute updates for the whole set of meta-windows during one epoch, since only a group of them are sampled.

For the baselines, we set the following parameters for all the experiments: batch size is 128, training epochs are 1000, and the patience for the early stopping is 50. In the experiments with baselines, two learning rates are considered. On the one side, the training learning rate that is used for finding the pre-trained parameters (\textit{Training LR}). On the other side, a different learning rate is considered for fine-tuning the models during test (\textit{Fine-tuning LR}). Both learning rates are chosen from a set of possibilities $\{0.01, 0.001, 0.0001\}$, by assessing the performance of the model with the pre-trained parameters and the performance after fine-tuning accordingly on the validation set. For VRADA, the training LR is fixed to 0.0003, following the value used by \cite{sanjay2017}. The weight decay for the fine-tuning is chosen from \{0, 0.5, 0.1, 0.01, 0.001, 0.0001\}.

\textbf{MAML} algorithm uses the same LSTM architecture as the baseline, and aims to find the best last-layers parameters so that it achieves a good performance in few updates for a new virtual task. It uses the following settings. The batch size (batch of meta-windows in algorithms \ref{alg:maml_for_tsr}) is 20. The meta-training epochs are set to 10000, similar to the baselines. The patience is 500, as the definition of meta-training epoch is slightly different to training epoch , and less data is considered in every epoch. We also include a noise level, to achieve meta-augmentation \cite{rajendran2020meta} by adding noise to the targets of the support sets, such that $y_{noise}=y+\epsilon, \epsilon \sim \{0, \textit{noise level}\}$. This is one of the proposed approaches \cite{rajendran2020meta} to make the model more robust against meta-overfitting. The grid for the hyper-parameter tuning via the meta-validation set are set as follows: meta-learning rate ($\beta$ in algorithm \ref{alg:maml_for_tsr}) \{0.0005, 0.00005\}, learning rate ($\alpha$) \{0.01, 0.001, 0.0001\} and noise level \{0, 0.01, 0.001\}. The hyperparameters are chosen so that it reduces the error for horizon 10 in all the experiments with MAML.

\textbf{MMAML} uses the same task network as MAML, a two layers LSTM, however the architecture involves other modules.  The encoder and the decoder are one-layer LSTMs with hidden size 128. The generator is a linear layer with 256 neurons, as it generates two vectors of parameters for the FiLM layer (each of dimensionality 128). We tune the same hyper-parameters as for MAML, but including the VRAE weight ($\lambda$ in equation  \ref{eq:mmaml_for_tsr}), considering the grid \{0.1,0.001, 0.0001\}. The final chosen configurations for the baselines and the proposed models are presented in the support material.

\subsection{Results}

We present the results of the proposed experiments in the table  \ref{table:results}. The bold font indicates the best results (lowest MAE) and the underlined font indicates the second best result. Additionally, the 95\% confidence interval is provided.

After the results, it is possible to observe that MMAML achieves overall good results, always better than transfer-learning on the same backbone (LSTM) and, most of the times, better than transfer learning even on more powerful models such as Resnet. Moreover, the performance, after fine-tuning, remains high for a long evaluation horizon when using MAML and MMAML as can be seen in the results on 10 horizons. It means they exhibit less temporal overfitting.

Another important insight is that MAML is less prone to overfitting after more gradient steps. By looking at the results on all the datasets, it is noticeable that on 10 gradient steps, the MAML algorithm still performs better in long horizons (10 Horizons) than MMAML. We hypothesize that this is because the parameter modulation from MMAML based on meta-features of a meta-window is more robust against overfitting so long as it is applied on meta-windows temporally close to the support meta-window (input of the modulation network), but might decrease the performance in temporally-far examples.

During the baselines evaluation, we notice that finding the best \textit{fine-tuning LR} given a pre-trained model is difficult, since the learning rate that works best for the validation set may not perform equally good for the meta-testing. They may overfit easily, as it happened on HR (1 Horizon, 1 gradient step), where the simplest baseline performed better. Here, our proposed methods set a clear advantage as the \textit{fine-tuning LR} is already fixed in the meta-training process. However, if the validation and test set are somewhat similar, for instance due to a small domain shift, the \textit{fine-tuning LR} tuned on the validation set may be suitable enough for the test set. Thus, a more powerful model will have an advantage over meta-learning approaches, as the results on BATTERY show.

\begin{table}[]
\caption{Results}

\setlength{\tabcolsep}{0.5em}
\resizebox{\textwidth}{!}{%
\begin{tabular}{ c  c  c  c  c  c }
\hline
\hline
\multirow{2}{*}{\textbf{Dataset}}                       & \multirow{2}{*}{\textbf{Model}} & \multicolumn{2}{c}{\textbf{1 Gradient Step}}         & \multicolumn{2}{c}{\textbf{10 Gradient Steps}}                       \\ 
                                                        &                                 & \textbf{1 Horizon}        & \textbf{10 Horizons}      & \textbf{1 Horizon}        & \multicolumn{1}{l}{\textbf{10 Horizons}} \\ \hline
\multirow{6}{*}{\textbf{POLLUTION}}                     & \textbf{Target Mean}            & 0.0465±0.0000             & 0.0495±0.0000             & 0.0465±0.0000             & 0.0495±0.0000                             \\ 
                                                        & \textbf{Resnet}                 & 0.0491±0.0074             & 0.0502±0.0062             & 0.0472±0.0047             & 0.0519±0.0057                             \\ 
                                                        & \textbf{VRADA}                  & 0.0444±0.0012             & 0.0428±0.0011             & 0.0438±0.0008             & 0.0429±0.0008                             \\ 
                                                        & \textbf{LSTM}                   & 0.0467±0.0009             & 0.0463±0.0012             & 0.0446±0.0006             & 0.0437±0.0005                             \\ 
                                                        & \textbf{MAML (ours)}                   & \underline{0.0421±0.0002} & \underline{0.0418±0.0003} & \underline{0.0423±0.0010} & \bftab{0.0416±0.0009}                     \\ 
                                                        & \textbf{MMAML (ours)}                  & \bftab{0.0410±0.0012}     & \bftab{0.0417±0.0007}     & \bftab{0.0411±0.0010}     & \underline{0.0420±0.0011}                 \\ \hline
\multirow{6}{*}{\textbf{HR}}                            & \textbf{Target Mean}            & \underline{0.0542±0.0000} & 0.0975±0.0000             & 0.0542±0.0000             & 0.0975±0.0000                             \\
                                                        & \textbf{Resnet}                 & 0.0670±0.0063             & 0.0817±0.0035             & 0.0625±0.0043             & 0.0734±0.0024                             \\ 
                                                        & \textbf{VRADA}                  & 0.0789±0.0066             & 0.0799±0.0060             & 0.0761±0.0084             & 0.1140±0.0071                             \\ 
                                                        & \textbf{LSTM}                   & 0.0673±0.0002             & \underline{0.0788±0.0006} & 0.0565±0.0004             & 0.0906±0.0010                             \\ 
                                                        & \textbf{MAML (ours)}                   & 0.0634±0.0018             & 0.0792±0.0029             & \underline{0.0511±0.0022} & \bftab{0.0711±0.0068}                     \\ 
                                                        & \textbf{MMAML (ours)}                  & \bftab{0.0448±0.0009}     & \bftab{0.0689±0.0015}     & \bftab{0.0507±0.0008}     & \underline{0.0729±0.0027}                 \\ \hline
\multirow{6}{*}{\textbf{BATTERY}} & \textbf{Target Mean}            & 0.0255±0.0000             & 0.0658±0.0000             & 0.0255±0.0000             & 0.0658±0.0000                             \\ 
                                 & \textbf{Resnet}                 & \bftab{0.0184±0.0024}     & \bftab{0.0141±0.0007}     & \bftab{0.0091±0.0007}     & 0.0160±0.0012                             \\ 
                                  & \textbf{VRADA}                  & 0.0352±0.0019             & 0.0309±0.0016             & 0.0967±0.0135             & 0.0995±0.0124                             \\ 
                                 & \textbf{LSTM}                   & 0.0407±0.0025             & 0.0417±0.0024             & 0.0195±0.0013             & 0.0217±0.0013                             \\ 
                                & \textbf{MAML (ours)}                   & 0.0243±0.0012             & 0.0170±0.0010             & \underline{0.0135±0.0006} & \bftab{0.0115±0.0004}                     \\ 
                                 & \textbf{MMAML (ours)}                  & \underline{0.0206±0.0021} & \underline{0.0154±0.0020} & 0.0156±0.0023             & \underline{0.0149±0.0022}                 \\ \hline \hline
\end{tabular}}
\label{table:results}
\end{table}

\begin{figure}[H]
\subfloat[]{\label{fig:hr_steps}{\includegraphics[width=0.5\textwidth]{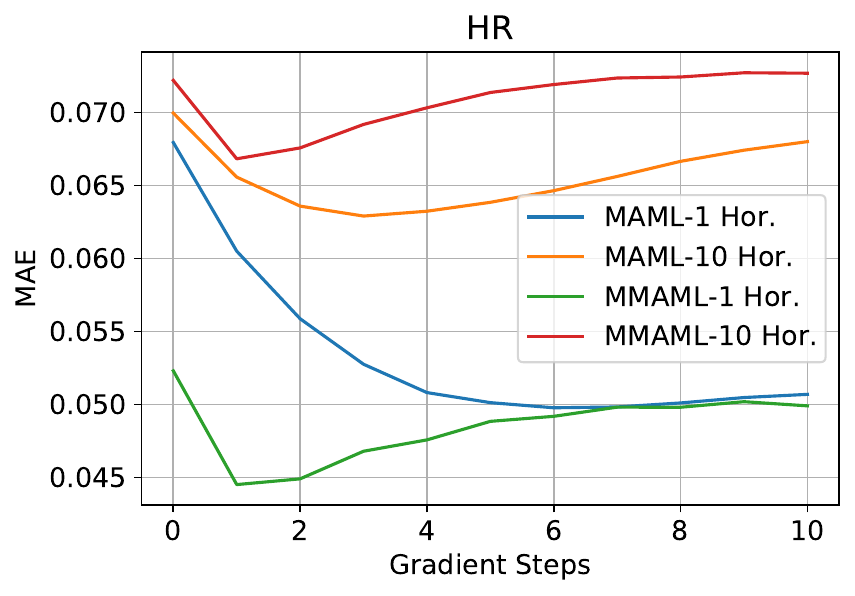}}}\hfill
\subfloat[]{\label{fig:hr_horizons}{\includegraphics[width=0.5\textwidth]{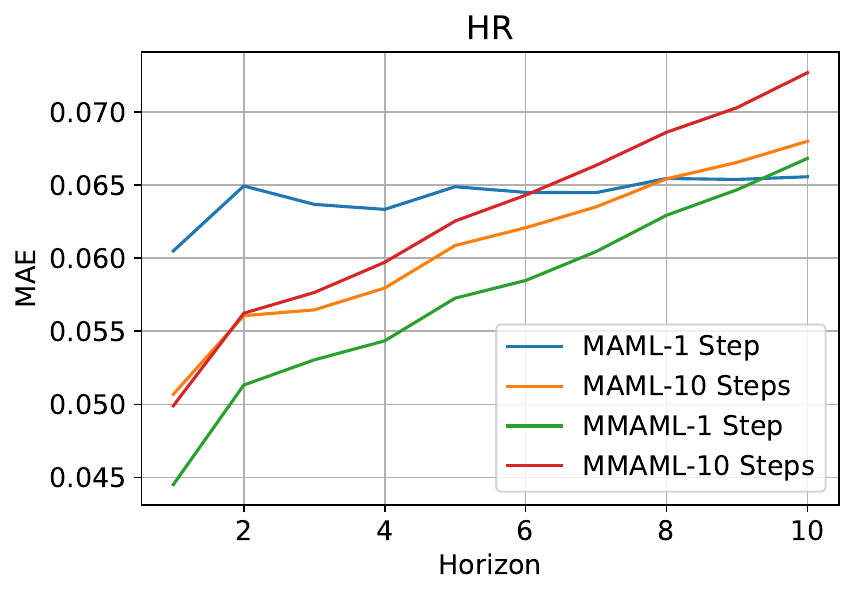}}} \hfill
\caption{Change in MAE while increasing the gradient steps during fine-tuning on meta-testing.}
\label{fig:hr_horizons_and_step}
\end{figure}

\subsection{Ablation studies}

We run some additional experiments to test the performance of our proposed algorithms under different configurations. Firstly, we would like to see how the error behaves when the models are fine-tuned beyond the number of gradient steps assumed during the meta-training. 
A look at figure \ref{fig:hr_horizons_and_step} makes possible to understand this and how the overfitting may arise on our proposed methods under different horizons. The lowest error is achieved after one gradient step even on long horizons, except for MAML on 1 horizon, where MAE keeps decreasing after several gradient updates (figure \ref{fig:hr_steps}). This shows that MAML is robust against overfitting on close horizons, after several gradient steps. However, when having more updates, MAML may overfit temporally, thus exhibiting bad performance in long horizons as the orange curve depicts in figure \ref{fig:hr_horizons}.

\begin{figure}[]
\subfloat[]{\label{fig:left_vrae}{\includegraphics[width=0.3\textwidth]{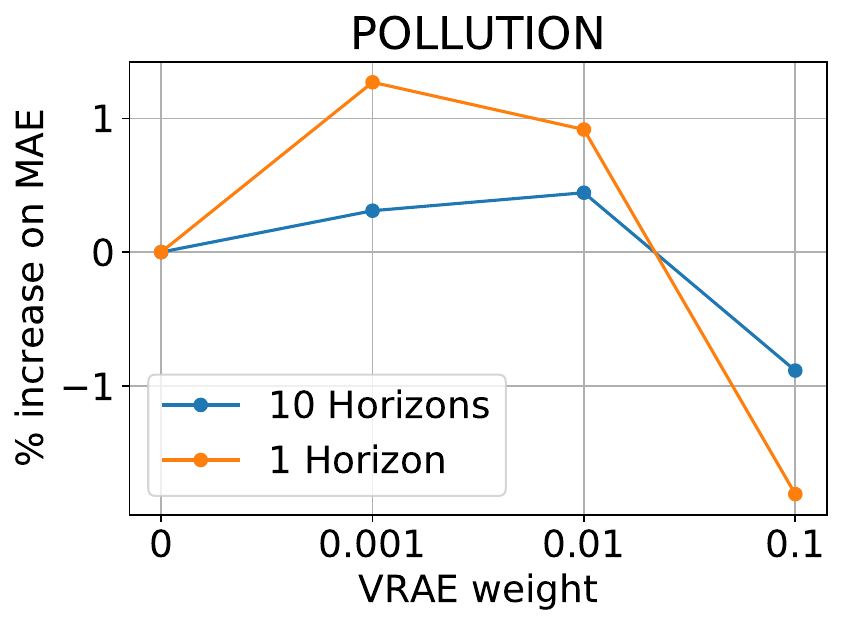}}}\hfill
\subfloat[]{\label{fig:mid_vrae}{\includegraphics[width=0.3\textwidth]{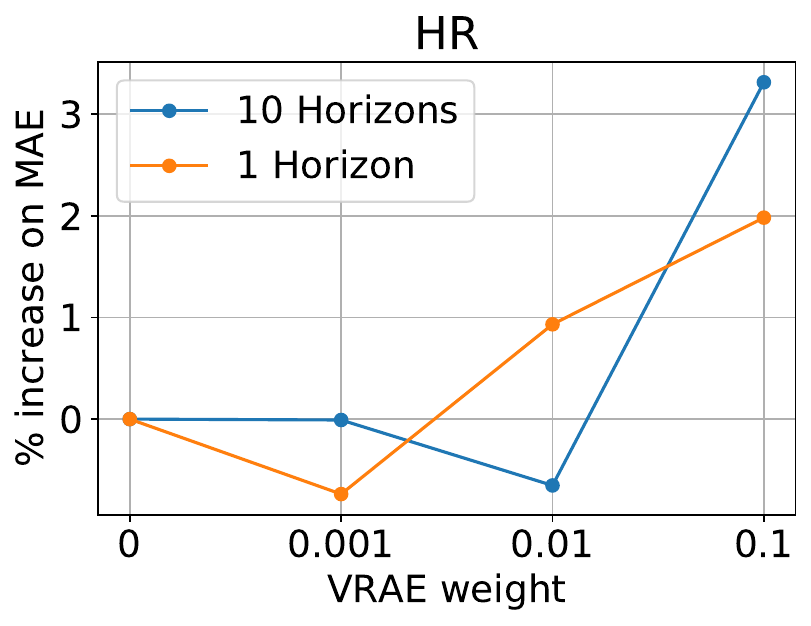}}} \hfill
\subfloat[]{\label{fig:right_vrae}{\includegraphics[width=0.3\textwidth]{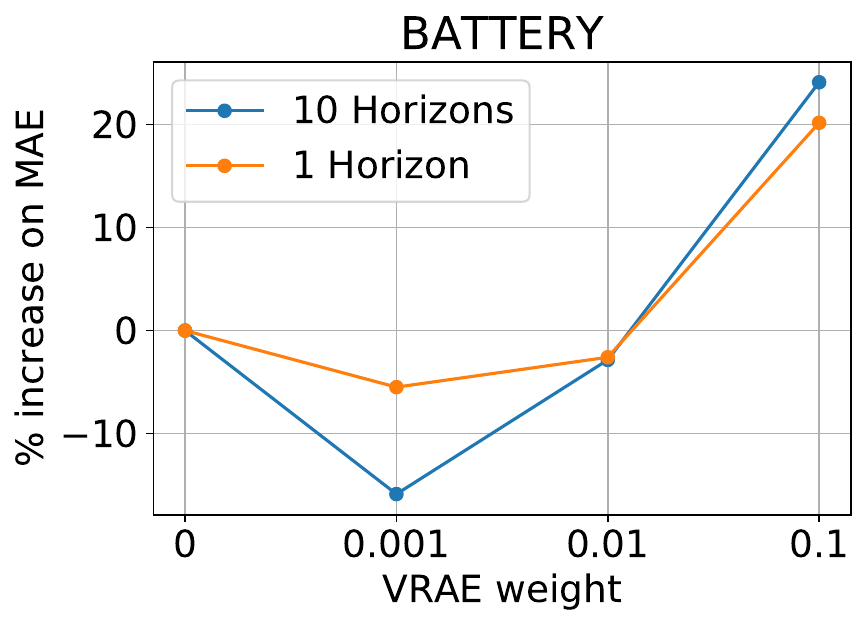}}} \hfill

\caption{Change in MAE with respect to VRAE. }
\label{fig:vrae_weight}
\end{figure}

The figures \ref{fig:left_vrae}, \ref{fig:mid_vrae} and \ref{fig:right_vrae} show that there is indeed an advantage of including the variaional loss in our formulation, as there is a decreased MAE when having values for the VRAE weight ($\lambda$) different from zero.

\section{Conclusion}

The present work introduces an extension of Meta-Agnostic Meta-Learning (MAML) and Multi-modal MAML (MMAML) to time series regression (TSR). We propose a design for the tasks such that we leverage the original, more common scenario of few but long time series available. The proposed design, which introduces the concept of "meta-window", makes possible to have more tasks available for meta-training. Through experiments, we show how this idea works on different datasets, allowing to achieve better performance than traditional methods such as transfer learning.  This is the first time to apply meta-learning for fast adaption on TSR, and it shows that it is possible to adapt to new TSR tasks with few data and within few iterations. For future work, we hypothesize that the application of the introduced ideas would have promising results in time series forecasting.

\newpage
\section*{Acknowledgements}
The research of Kiran Madhusudhanan is co-funded by the industry project \href{https://www.ismll.uni-hildesheim.de/projekte/ecosphere_en.html} {"IIP-Ecosphere: Next Level Ecosphere for Intelligent Industrial Production"}. Sebastian Pineda Arango would also like to thank Volkswagen AG who funded his intership in order to carry out this research.
%
%
%
\bibliographystyle{splncs04}
\bibliography{bibliography1}
%


\end{document}